%% file: pagination.tex
\documentclass{sig-alternate}

\usepackage{color}
\usepackage{xspace}
\usepackage[table]{xcolor}
\usepackage{graphicx}
\usepackage{caption}
\usepackage{subcaption}
\usepackage{multirow}
\usepackage{longtable}
\usepackage{supertabular}
\usepackage{sparklines}
\usepackage{rotating}
\usepackage{pgfplots}
\usepackage{booktabs}
\usepackage{tablefootnote}
\usepackage{comment}

\newcommand{\hide}[1]{} 

\definecolor{tableShade}{HTML}{F0F0F0}

\def \sOne {First Sentence\xspace}
\def \sTwo {Second Sentence\xspace}
\def \pOne {First Paragraph\xspace}
\def \tw {Twenty Percent\xspace}
\def \na {Novelty Article\xspace}
\def \nc {Novelty Corpus\xspace}
\def \ka {SLM Article\xspace}
\def \kc {SLM Corpus\xspace}

\def \sparklineFootnote {Ratings based on a 7-Point scale:\\ (1) ``missing information'' (4) ``balanced'' and (7) ``excess text.''}

\def \noveltySize {0.19}
\def \slmSize {.23}
\def \slmPointsColor {black}
\def \slmLineColor {gray}
\def \slmXAxis { Sentence Number}
\def \slmYAxis { KL Divergence}
\def \slmYYAxis { Std Deviation}
\def \smlFontSize {\tiny}
\def \markSize {0.4pt}
\def \slmMaxY {4}

\begin{document}
%
\conferenceinfo{WOODSTOCK}{'97 El Paso, Texas USA}

\title{Pagination: It's what you say, not how long it takes to say it}
%
%
%
%
%

\numberofauthors{3} 
%
\author{
%
%
\alignauthor
Joshua Hailpern\\
       \affaddr{HP Labs}\\
       \affaddr{Palo Alto, CA, USA}\\
       \email{joshua.hailpern@hp.com}
\alignauthor
Niranjan Damera Venkata\\
       \affaddr{HP Labs}\\
       \affaddr{Chennai, Tamil Nadu, India}\\
       \email{niranjan.damera-venkata@hp.com}
\alignauthor
Marina Danilevsky\\
       \affaddr{University of Illinois}\\
       \affaddr{Urbana, IL, USA}\\
       \email{danilev1@illinois.edu}
}


\maketitle
\begin{abstract}
\input{./section_abstract}
\end{abstract}




\input{./section_intro}

\input{./section_relatedwork}

\input{./section_scopeterms}

\input{./section_lowerbound}

\input{./section_dataset}

\input{./section_performance}

\input{./section_gold}

\input{./section_conclusion}



\hide{
\appendix
\section*{APPENDIX}
\setcounter{section}{0}

\section{Finding Inflection Points}\label{appx:inflection}
\begin{enumerate}
\item Take the log of the number of sentences (convert to linear)
\item Normalize the weights from 0.0 to 1.0
\item Create a linear regression with the x-axis to be log of the number of sentences, and y-axis to be normalized sentence weights 
\item Use the x-axis coefficient (e to the coffient) as the inflection point
\end{enumerate}
}

%
\bibliographystyle{abbrv}

\bibliography{library}  
%
%
\end{document}

%% file: section_abstract.tex

Pagination - the process of determining where to break an article across pages in a multi-article layout - is a common layout challenge for most commercially printed newspapers and magazines. To date, no one has created an algorithm that determines a minimal pagination break point based on the content of the article. Existing approaches for automatic multi-article layout focus exclusively on maximizing content (number of articles) and optimizing aesthetic presentation (e.g., spacing between articles). However, disregarding the semantic information within the article can lead to overly aggressive cutting, thereby eliminating key content and potentially confusing the reader, or setting too generous of a break point, thereby leaving in superfluous content and making automatic layout more difficult. This is one of the remaining challenges on the path from manual layouts to fully automated processes that still ensure article content quality. In this work, we present a new approach to calculating a document minimal break point for the task of pagination. Our approach uses a statistical language model to predict minimal break points based on the semantic content of an article. We then compare 4 novel candidate approaches, and 4 baselines (currently in use by layout algorithms). Results from this experiment show that one of our approaches strongly outperforms the baselines and alternatives. Results from a second study suggest that humans are not able to agree on a single ``best'' break point.  Therefore, this work shows that a semantic-based lower bound break point prediction is necessary for ideal automated document synthesis within a real-world context.


%% file: section_intro.tex
\section{Introduction}

Traditional document composition is an iterative process involving copy editors and professional publication designers who work in concert to make decisions on what content to include and how to format it for aesthetic presentation. Within the context of newspapers or magazines, each page can contain multiple articles. However, space constraints dictate that the full text of any (or all) given articles can not be presented on a single page. Articles are therefore broken or paginated across a paper. The first portion of an article is presented upfront, while the full text is presented later on, should the reader wish to consume more. Copy editors and designers work to strike a balance between presenting enough content on this front page so readers can understand the article, while maximizing the space constraints and aesthetic presentation of the page as a whole.

Automated document composition algorithms have been introduced to automate this largely manual workflow. Current approaches to automated document synthesis/layout focus extensively on content and aesthetic maximization \cite{Hurst:2009et,Jacobs:2003hl}. These optimization algorithms work with a series of rules and constrains to maximize the presentation within the spatial and visual constraints.

While these approaches have largely been successful at addressing issues of presentation, they do not take into account the content of the articles. This can result in break points being set too aggressively, resulting in missing key content and a potential misrepresentation of the article content, or too late in the article, potentially lowering the flexibility of layout algorithms by requiring more content to be placed on the front page. To combat these problems, many of these layout algorithms enforce minimal content requirements. However, these rules are arbitrary, consisting of static or relative lower bounds (e.g., at least 2 sentences, or at least 20\% of the article). Given that these rules are not related to the {\em actual} article, key content can still be left out, or excess text may still be forced to be included.


The primary contribution of this work is an algorithm that predicts a lower-bound break point for news article pagination based upon the semantic information contained within the article. Our approach not only outperforms existing solutions, but to the authors' knowledge, is also the first such semantic based pagination algorithm.

In this paper we develop four candidate break point algorithms, all of which are based on the semantic information of an article to be paginated. These four approaches are then directly compared against four baselines currently used in both print and digital media. To facilitate these comparisons, we conduct an extensive Mechanical Turk study across 7 subject areas (Sports, US News, Entertainment, etc). Results from this experiment show that one of our approaches strongly outperforms the baselines and alternatives. We also investigate whether an `ideal' break point could be found for paginating news articles. Results from a large CrowdFlower study strongly suggest that humans can not agree on a single best break point. Therefore, this work shows that a semantic-based lower bound break point prediction is necessary for ideal for automated document synthesis within a real-world context.

%% file: section_relatedwork.tex
\section{Related Work}

The majority of work on pagination - the need to split a document or series of documents over multiple pages - focuses on content maximization and aesthetic presentation. Such approaches aim to create `high-quality documents,' defined as documents `without unwanted empty areas'  \cite{Ciancarini:2012tw}. Within the context of a single document, such as a book or a paper, pagination impacts the layout relationship of text content to related figures or tables \cite{BruggemannKlein:2003we}. 

However, the problem complexity greatly increases for automated document layout of newspapers or magazines\cite{Giannetti:2008fg,Hurst:2009et,Jacobs:2003hl}. In addition to optimizing the placement of figures within text, a given page can also contain content from multiple articles, and all content must fit within a predetermined page count. When determining how to break up articles across multiple pages, the preferred approach is to use constraint-based layout models, in which layout specifications are described by linear constraints imposed on the items within a layout \cite{Badros:2001is,Jamil:2012tw,Lutteroth:2008tu,Scoditti:2009hw,Zeidler:2012ct}. The focus of these models is on the composition quality \cite{Ahmadullin:2013dj,DameraVenkata:2011hv}, and they in no way account for the semantic information contained within the articles themselves.

\subsection{Automatic Summarization}

A similar problem to pagination is that of single document summarization - the process by which the text of an article is reduced either by extraction (lifting sentences from the original text)\cite{Mihalcea:2004wf} or abstraction (using natural language processing techniques to generate new sentences)\cite{Fiszman:2004wo}. While these approaches are not used by layout or copy editors, they are relevant to the construction and evaluation of our semantic-based break point techniques.

Corpus summarization uses a large collection of documents to build a model of the topics being discussed (e.g. topic modeling \cite{Blei:2003tn,Chua:2013td}, SumBasic\cite{Haghighi:2009wl}, KLSum\cite{Nenkova:2005vd}) or opinions rendered (e.g. Opinion Mining \cite{Hailpern:2013cv}). Corpus summarization approaches rely upon a large body of documents (e.g., a collection of tweets \cite{Chua:2013td}) from which patterns about the `whole' can be derived, and are generally easier and more powerful because they have more data from which to draw summaries. In contrast, single document summarization \cite{Lin:1997tf,Katragadda:2009vu} utilizes only one document to create a summary.  Within single document summarization, most algorithms are designed to summarize long (e.g. book)\cite{Yang:2008tg}, well structured (e.g. chapters or sections) text\cite{Edmundson:1969dt,Seki:2004vn,Yang:2003th}, thus maximizing the amount of text and structural cues from which to derive summaries. The most notable exceptions to single unstructured documents are TextRank\cite{Mihalcea:2004wf} and LexRank\cite{Erkan:2004vn}. Both of these algorithms use a simple graph-based approach, treating each sentence as a node. The summary sentence of the document is calculated by finding the centroid of the graph based on a distance vector. There has been some work specifically with summarizing news articles by extracting the most important facts from the article \cite{Kastner:2009tc}. Finally, it has shown that using the first n sentences of a news article as the article's summary performs very well \cite{Nenkova:2005wx}, which adds motivation to our end goal.

\subsection{Document Sub-Topic Segmentation}

Another tangential, though different, problem space is that of document sub-topic segmentation, most notably TextTiling \cite{Hearst:1997wq,Pevzner:2002uh}. The main focus of Sub-Topic Segmentation is to divide a document into subtopics, or sections. These sections are not ranked, as to importance or quality, but strive to focus on conceptual shifts in the content being discussed. TexTiling is a straightforward and simplistic \cite{Beeferman:1999km} method the examines lexical co-occurred of terms between phrases in a document, and identifying sharp changes as breaks or dividers between subtopics. However, because these approaches lack of ranking or quality of each sub-topic shift, their direct application to the pagination problem are limited and a direct comparison of TextTiling with this problem is difficult. However, this work does suggest techniques or broad approaches we can build upon; detecting breaks based on semantic change variation across a document relative to a mean change, using occurrent of vocabulary to signify information/theme \cite{Hearst:1997wq}. 

%% file: section_scopeterms.tex
\section{Scoping \& Key Terms}

The goal of this work is to create and test a novel algorithm that can use a news article's semantics (rather than layout/spacing optimization) to determine a lower bound where to break an article for pagination. This serves two key purposes: first, if an article is broken too early, the reader may miss key information and potentially be mislead about the article's content; second, if too much of an article's content is forced to be shown, the page layout may not be as flexible because more text would be required to be presented therefore, there may not be space to present as many articles on the front page.

While there are many applications of this technique, this work specifically looks at news articles. Our novel approaches draw on and compare with pagination techniques both in print and digital presentation.

In this context, we define the following terms:

{\quote

{\bf Topic:} Refers to the use in the english vernacular, such as the topic of a news article (e.g. a specific bomb that goes off in a specific country on a specific day), rather as it is used in the context of Topic Modeling approaches. 

{\bf Article:} A piece of written text about a specific topic (e.g. a specific new york times article about a bomb that went off in a country).

{\bf Break Point:} The location at which you `break' or stop an article, such that the content before the break point is displayed on the current page, and the content after the break point goes on a different page, or after a `read more' link.

{\bf Lower Bound Break Point:} The location within a news article denoting that the reader would have, by this time, gained a general understanding about the topic being discussed, but has not yet seen all the details or nuances. Breaking before this lower bound is likely to cause the reader to miss a key concept or aspect of the article, and risk being mislead about the content. However, if more space is available, a layout algorithm can choose to include more text from a given article.

{\bf Pagination:} the task of determining where to break an article across a multi-article, multi-page layout.

{\bf Corpus:} A collection of articles on the same topic (e.g. 100 news articles about a specific bombing event).

{\bf Subject:} An overarching thematic grouping of articles or corpuses (e.g. Sports, Politics)

{\bf Document:} Any set of sentences, which could be a single article, a subset of an article, multiple articles, a Corpus or Subject, etc.

{\bf Semantics:} Refers to the english definition of ``semantics,'' focusing on the {\em meaning} of words or phrases.

}

It should be noted that there is an important distinction between pagination (the focus of this paper) and truncation (not discussed in this paper). Pagination, as used in this context, implies that after the break point, a reader will see more of the given article, usually on a subsequent page. On the contrary, when an article is truncated, the portion of the article which follows the break point will never be presented to the reader in any form. We do {\em not} deal with truncation tasks in this work. Further, this work does not perform document summarization, or sub-topic identification. These approaches can be used to inform our design, but the functionally address and solve a different problem and their output is not directly comparable to the output of pagination.

In this paper, we first describe the algorithms developed for content-aware pagination, as well as the baselines for comparison. Next, we present our experiment comparing the quality of the suggested break points based on article content. Finally, we briefly explore the challenges of predicting an `ideal' break point in a document.

%% file: section_lowerbound.tex
\input{./noveltyFigure}

\section{Lower Bound Algorithms}

The primary goal of this works is to develop and compare algorithms that can predict a lower bound break point for pagination of news articles. This section details the four content-agnostic algorithms that are used as `current state' baselines, as well as our novel predictive algorithms that directly leverage the semantic content of the articles.

\subsection{Four Baselines}

To facilitate a comparison, we leveraged four baselines that are commonly used as lower-bounds for pagination: the first sentence, the first two sentences, the first paragraph, and the first twenty percent of the article.

\subsubsection{One \& Two Sentences}
Commonly used in online news websites (e.g. Google News, NBCNews.com), the first or first two sentences of a news article are used to `preview' the full text. Thus, pagination occurs after these sentences, and the reader can consume the full text by clicking a link\footnote{It should be noted, that some online websites (e.g. New York Times) create custom one or two sentence  previews, rather than drawing them directly from the original text. This research does not examine the quality of these custom hand-crafted previews, which are more akin to single-document summaries.}. We refer to these two techniques as {\bf One Sentence} and {\bf Two Sentences}, respectively.

\subsubsection{One Paragraph}
While writing, a structurally delimiter between semantic concepts is often the paragraph break. Thus, the first paragraph of an article can be thought of as the first digestible nugget of an article's content and may make a natural break point. This is used at Wall Street Journal's website (wsj.com) for non-subscribers who seek to access subscriber only articles. We will refer to this as {\bf One Paragraph}.

\subsubsection{Twenty Percent Rounded Up to Nearest Paragraph}
When constructing printable newspapers, many sources use an article length dependent approach. Minimal breakpoints are calculated by the first 20\% of a document (measured by character count), then rounded up to the end of the current paragraph. This approach assumes that within the first fifth of a document, key concepts have been presented to the reader, and are roughly bounded by paragraph delimitation. We refer to this technique as {\bf Twenty Percent}. 

\subsection{Keyword Novelty}

The first approach we develop is predicated on the idea that as a reader traverses through a document, he or she is exposed to key concepts/words. First exposure to a given key word is enough to make the reader aware of that subject and that it is relevant to the article\footnote{Exactly how the given subject is related to the full text may not be clear until later in the article.}. To this end, the Keyword Novelty approach attempts to find when the reader has been exposed to `enough' of these key words that he or she would have a general understanding of the article.

\input{./slmFigure}

\subsubsection{Calculating Keyword Novelty}

The first step is determining what are the key words in a document. Following standard IR techniques, we limit a document's text to information-heavy words (nouns), and remove any pluralization through lemmatization. However, not all of the remaining keywords are equally relevant to the document in question. Commonly, term frequency (TF) can be used as a proxy for keyword relevance. However, TF  is generally not robust on shorter or sparse data, such as a single newspaper article.

An alternative technique to discovering keyword important is to use Singular Value Decomposition (SVD). SVD is able to filter out the noise in relatively small or sparse data, and is often used for dimensionality reduction. To repurpose SVD to calculate word weight, we represent each sentence as a row in a sentence-word occurrence matrix encompassing $m$ sentences and $n$ unique words, which we will refer to as {\bf M} (which can be constructed in O(m)).  

SVD decomposes the $m \times n$ matrix {\bf M} into a product of three matrices: {\bf M = U $\Sigma$ V*}. $\Sigma$ is a diagonal matrix whose values on the diagonal, referred to as $\sigma_i$, are the singular values of M. By identifying the four largest $\sigma_i$ values, which we refer to as $\lambda_1-\lambda_4$, we are able to take the corresponding top eigenvector columns of V (which is the conjugate transpose of V*), which we refer to as $\xi_1-\xi_4$. Note that each entry in each of these vectors $\xi_1-\xi_4$ corresponds to a unique word in {\bf M}.

We then create $\xi'$, a master eigenvector calculated as the weighted average of $\xi_1-\xi_4$, weighted by $\lambda_1-\lambda_4$: 

\begin{equation} 
\xi' = \frac{1}{4} \sum_{i=1}^{4} \lambda_i \xi_i 
\end{equation}

Thus, $\xi'$ is a vector in which each entry represents a unique word, and the value can be interpreted as the `centrality' of the word to the given document \cite{Yang:2008tg}. 

Once we have the keyword weights,\footnote{To further reduce noise, we use only the top 500 words from SVD.} we iterate over each sentence in a given article. The `value' of a sentence is the sum of all of the unique keywords' weights seen up to (and including) that sentence. Thus, the weight of a given sentence is a cumulative sum and each keyword only contributes to the overall sum on its first occurrence.

When plotted, the resulting sequence generally fits a logarithmic curve (see Figure \ref{fig:noveltyCurves}), allowing us to consider the inflection point\hide{\footnote{A point on a curve where the curvature changes sign.}} as a `lower bound' break point.\footnote{To find the inflection point:
\begin{enumerate}
\item {\em Take the log of the number of sentences (convert to linear)}
\item {\em Normalize the weights from 0.0 to 1.0}
\item {\em Create a linear regression with the x-axis to be log of the number of sentences, and y-axis to be normalized sentence weights} 
\item {\em Use the x-axis coefficient (e to the coefficient) as the inflection point}
\end{enumerate}}
The inflection point is an ideal point to paginate in that it is the location where the amount of text needed (space) is increasing more than the amount of new concepts (keywords).

\subsubsection{Article vs. Corpus Novelty}
Based on the keyword novelty metric, we created two variations:
\begin{itemize}
\item {\bf Article Keyword Novelty:} SVD weights are based on the individual article in question
\item {\bf Corpus Keyword Novelty:} SVD weights are based on all articles in a corpus treated as a single document
\end{itemize}
The corpus approach takes a more holistic view of the topic being discussed, whereas the article approach is more sensitive to the specific issues being addressed in a specific article.

\subsection{Statistical Language Modeling (SLM)}

The second approach we present is based on the idea that there is a probabilistic distribution of words (and their frequencies) that are an `ideal' we strive to mimic (e.g. the full text of the article, or the distribution of words in a corpus). We use SLM to create a model of the `ideal' document, and, for a given portion of an article being made visible to the reader, we use an information theoretic measure to discover how closely the model of that article portion comes to the `ideal' model of the entire document. 

An added benefit of the SLM approach is the ability to smooth the keyword frequencies that are to common to the broad subject (in this work we use Dirichlet Prior Smoothing, which has been shown to be effective \cite{Zhai:2009tm}).

\subsubsection{Calculating SLM}

As in Keyword Novelty, we pre-filter the text in each article to only contain the lemmatized high-information (noun) words. 


In order to explain our use of SLM, consider {\bf {\em S}} to be the set of all subjects {\small (s$_{0}$ $\ldots$ s$_{g}$)} in our dataset. We work with one subject at a time, which we will refer to as $s_i$.

Let {\bf {\em D}} be the set of all articles {\small (d$_{0}$ $\ldots$ d$_{b}$)} in the given subject s$_{i}$ and {\bf {\em W}} to be the set of all unique words  {\small (w$_{0}$ $\ldots$ w$_{h}$)} in s$_{i}$. 

Denote the frequency of a given word {\em w$_{j}$} in a given document {\em d$_{k}$} as {\em f(w$_{j}$$|$d$_{k}$)}.  Then the total count of all words in {\em d$_{k}$} is calculated as:

{ \begin{equation}
\textstyle  T(d_{k}) = \sum\limits_{j=0}^{h} f(w_{j}|d_{k}) 
 \label{eq:total}
\end{equation}}

and the probability of of a given word  (w$_{j}$) in {\em d$_{k}$} is:

\begin{equation}
\textstyle  p(w_{j}|d_{k}) = \frac{f(w_{j}|d_{k})}{T(d_{k})}
 \label{eq:probDoc}
\end{equation}

This therefore allows us to calculate the probability of a word in a document, using Dirichlet Prior smoothing \cite{Zhai:2009tm}:
\begin{equation}
\textstyle  q(w_{j}|d_{k}) = \frac{f(w_{j}|d_{k}) + \mu*p(w_{j}|s_{i})}{T(d_{k}) + \mu}
 \label{eq:probSmooth}
\end{equation}

where $p(w_j | s_i)$ is the occurrence probability of the word $w_j$ in the entire subject $s_i$:

\begin{equation}
p(w_j | s_i) = \frac{\sum_{d_k \in s_i} f(w_j | d_k)}{\sum_{d_k \in s_i} T (d_k)}
 \label{eq:probSubj}
\end{equation}

and where the smoothing constant $\mu$ is estimated using \cite{Seo:2010gv}:
\begin{equation}\begin{split}
 \begin{gathered}
m_{w_{j}} = p(w_{j}|s_{i})\\
B_{w_{j}} = \sum\limits_{d_{k} \in s_{i}}^{}( ( \frac{f(w_{j}|d_{k})}{T(d_{k})}- m_{w_{j}})^2 )\\
\mu = \frac{\sum\limits_{w_j \in W}^{} \frac{B_{j}}{m_{w_{j}} * (1-m_{w_{j}})}}{\sum\limits_{w_j \in W}^{} \frac{B_{j}^2}{m_{w_{j}}^2 * (1-m_{w_{j}})^2}}
\end{gathered}
\label{eq:ranking}
\end{split}\end{equation}


As we traverse an article sentence by sentence, we redefine the variable {\bf {\em N}} to refer to the subset of sentences in a given article we have seen. Thus, we calculate the probability of a word in N, using Dirichlet Prior smoothing \cite{Zhai:2009tm} as:
\begin{equation}
\textstyle  q(w_{j}|N) = \frac{f(w_{j}|N) + \mu*p(w_{j}|s_{i})}{T(N) + \mu}
 \label{eq:probSmoothTest}
\end{equation}


To compare each successive test SLM to the Ideal document, we use the KL-Divergence metric (smaller is better):
\begin{equation}
KLDivergence = \sum\limits_{w_{j} \in d_k}^{} (ln(\frac{q(w_{j}|d_{k})}{q(w_{j}|N)}))*q(w_{j}|d_{k})
 \label{eq:kldivergence}
\end{equation}
Thus for each sentence in an article, we can calculate the KLDivergence score, returning another set of distributions. When plotted (x-axis=sentence, y-axis=KLDivergence), these distributions appear relatively linear, with occasional `jumps' when the models get closer/further apart (for article and corpus respectively). We can detect the first of these `jumps,'  and use it as our break point. To find a jump, we perform the following steps:

{
\begin{quote}
\begin{enumerate}
\item Calculate the delta in KL-Divergence between having seen $N$ sentences and $N+1$ sentences, for the entire document.
\item Determine the mean and standard deviation of this set of delta values.
\item if the delta in KL-Divergence between having seen $N$ and $N+1$ sentences is greater than or equal to 2 standard deviations away from the mean, there a `jump' after the $N^{th}$ sentence.\footnote{We ignore the change between the first and second sentence since, due to many articles having `low' value first sentences, having almost any content in the second sentence creates a large delta.}
\end{enumerate}
\end{quote}}

\subsubsection{Variations}
As with the keyword novelty metric, we created two variations:
\begin{itemize}
\item {\bf Article SLM:} The ideal SLM is generated based upon the full text of the specific article
\item {\bf Corpus SLM:} The ideal SLM is generated based on all articles in the Corpus treated as a single document
\end{itemize}
The corpus approach takes a more holistic view of the topic being discussed, whereas the article approach is more sensitive to the specific issues and phrasing being addressed in a specific article.

%% file: noveltyFigure.tex
\begin{figure*}[!htb]
 \centering
  \begin{subfigure}[b]{\noveltySize\textwidth}
 \centering
 \includegraphics[width=\textwidth]{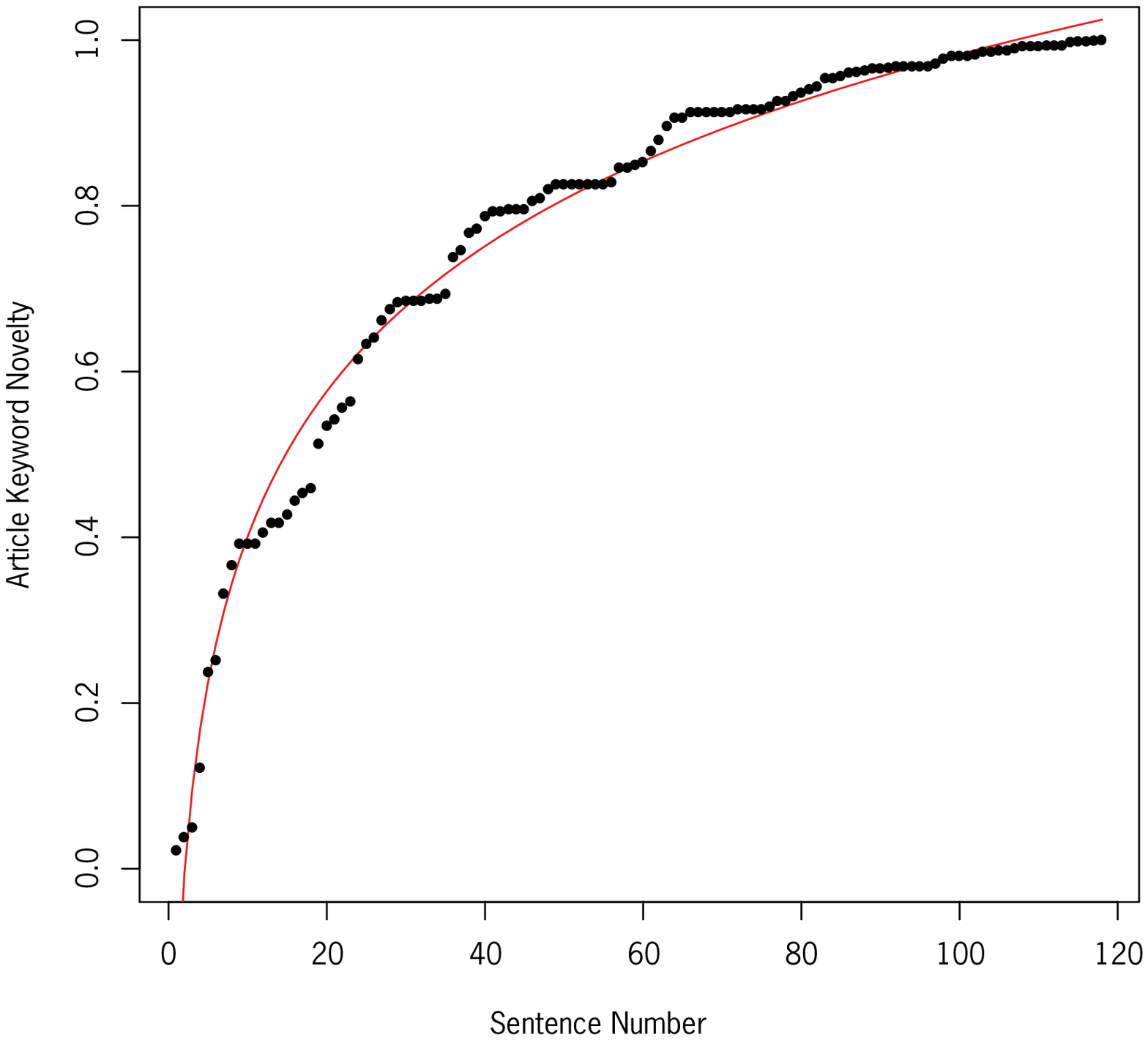}
 \caption{Buisness}
 \label{fig:splash}
 \end{subfigure}%
 ~ 
 \begin{subfigure}[b]{\noveltySize\textwidth}
 \centering
 \includegraphics[width=\textwidth]{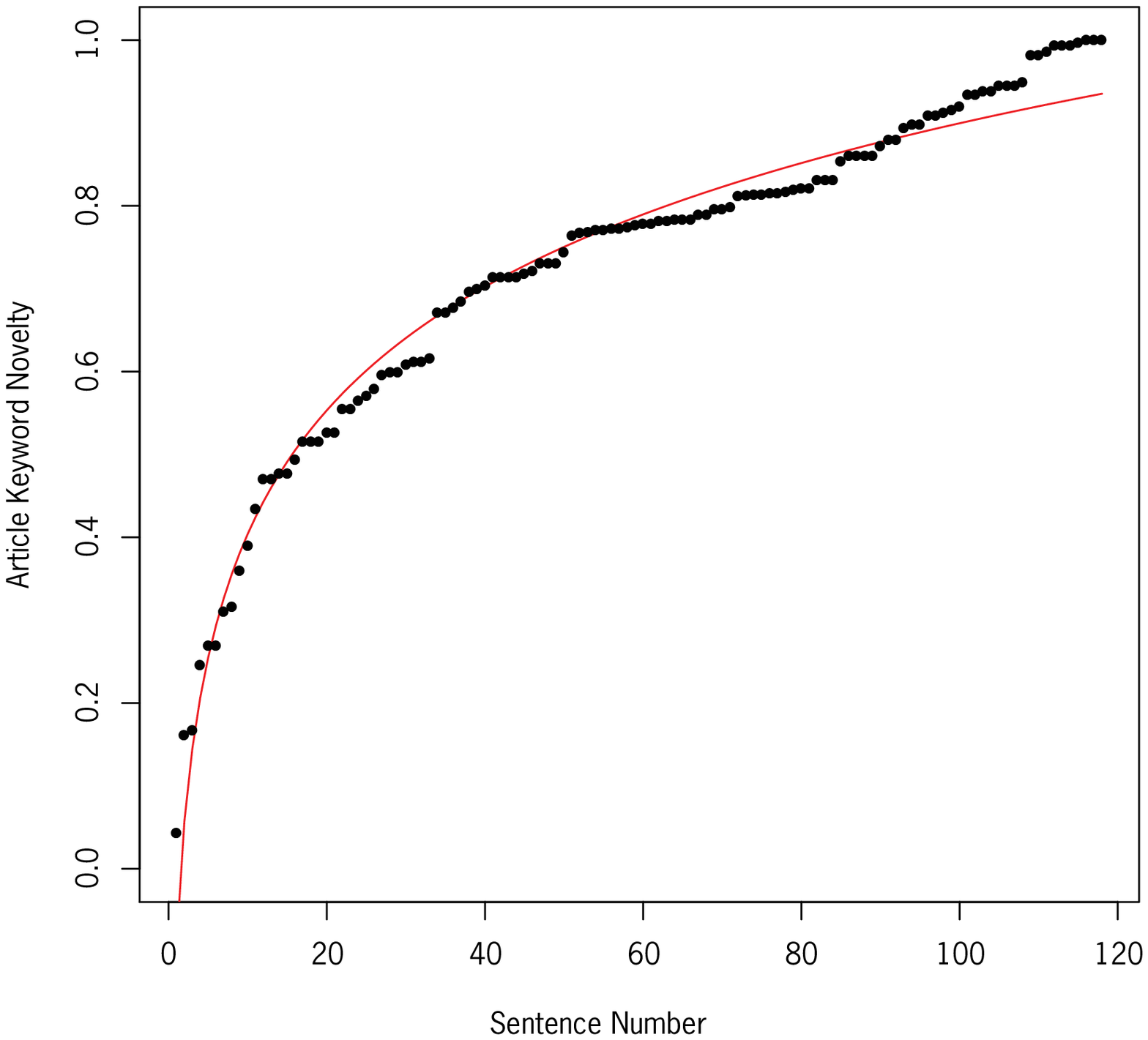}
 \caption{Entertainment}
 \label{fig:relMost}
 \end{subfigure}%
 ~ 
 \begin{subfigure}[b]{\noveltySize\textwidth}
 \centering
 \includegraphics[width=\textwidth]{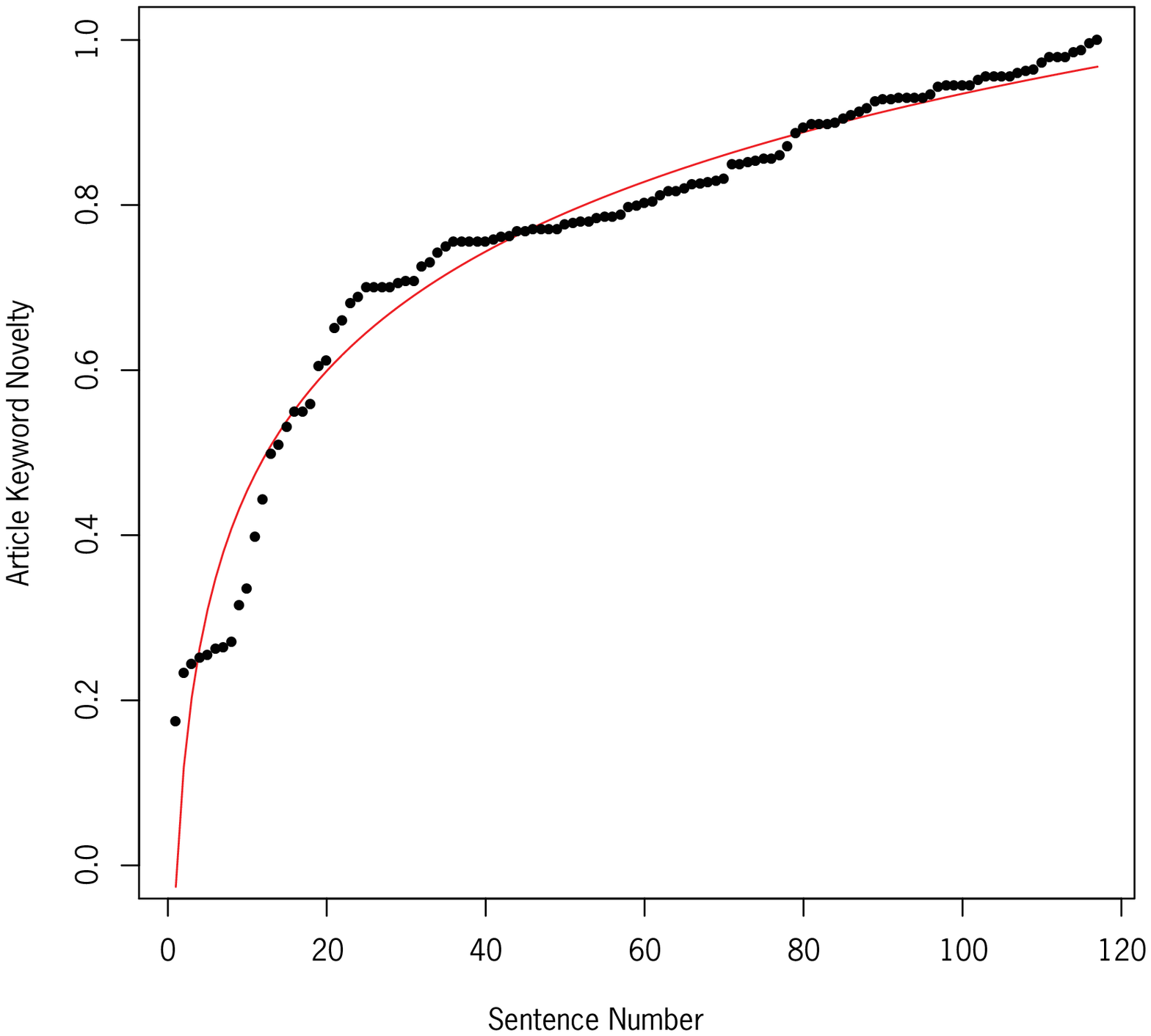}
 \caption{Politics}
 \label{fig:aliMost}
 \end{subfigure}
 ~ 
 \begin{subfigure}[b]{\noveltySize\textwidth}
 \centering
 \includegraphics[width=\textwidth]{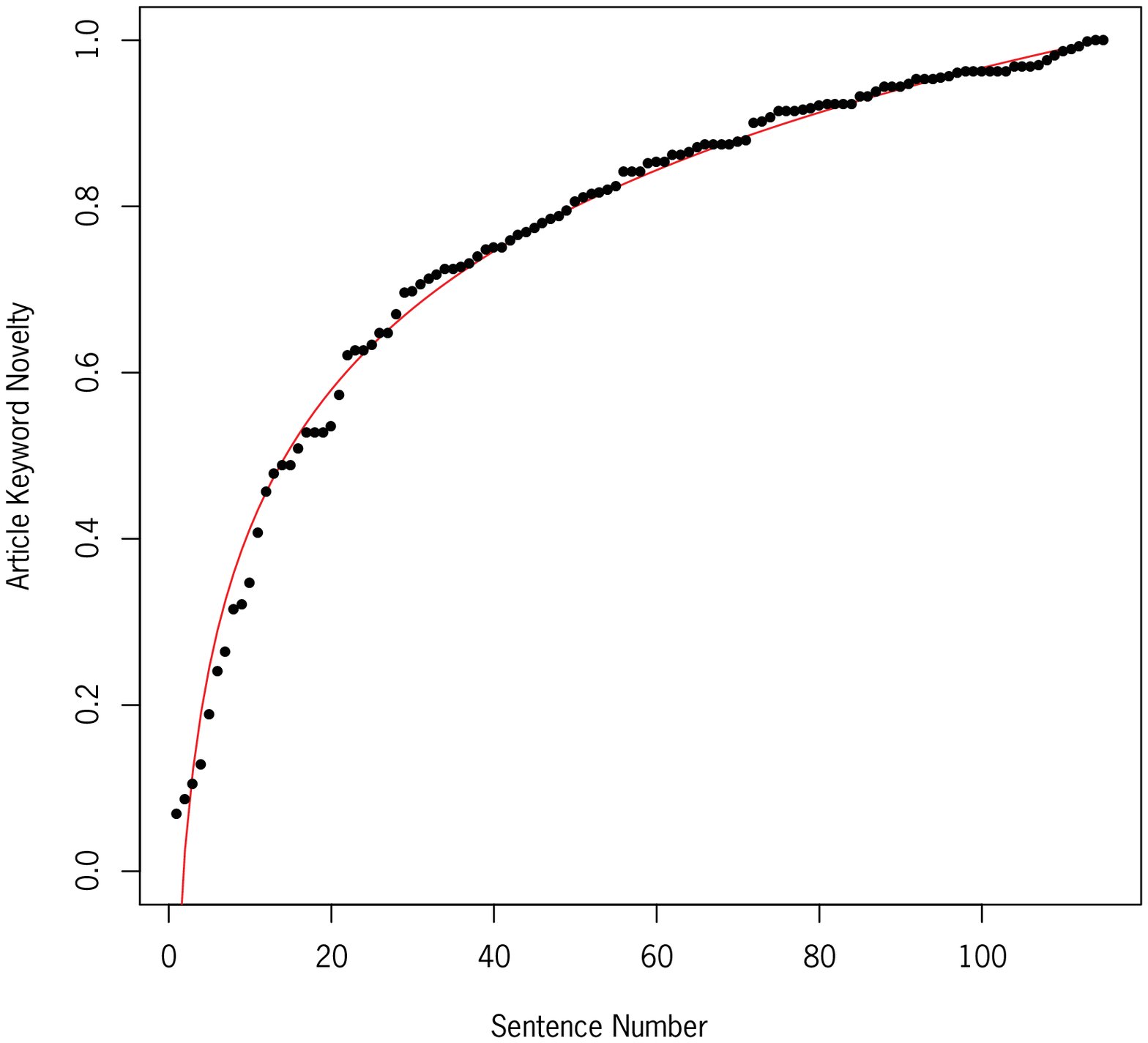}
 \caption{Sports}
 \label{fig:divMost}
 \end{subfigure}\\
 \begin{subfigure}[b]{\noveltySize\textwidth}
 \centering
 \includegraphics[width=\textwidth]{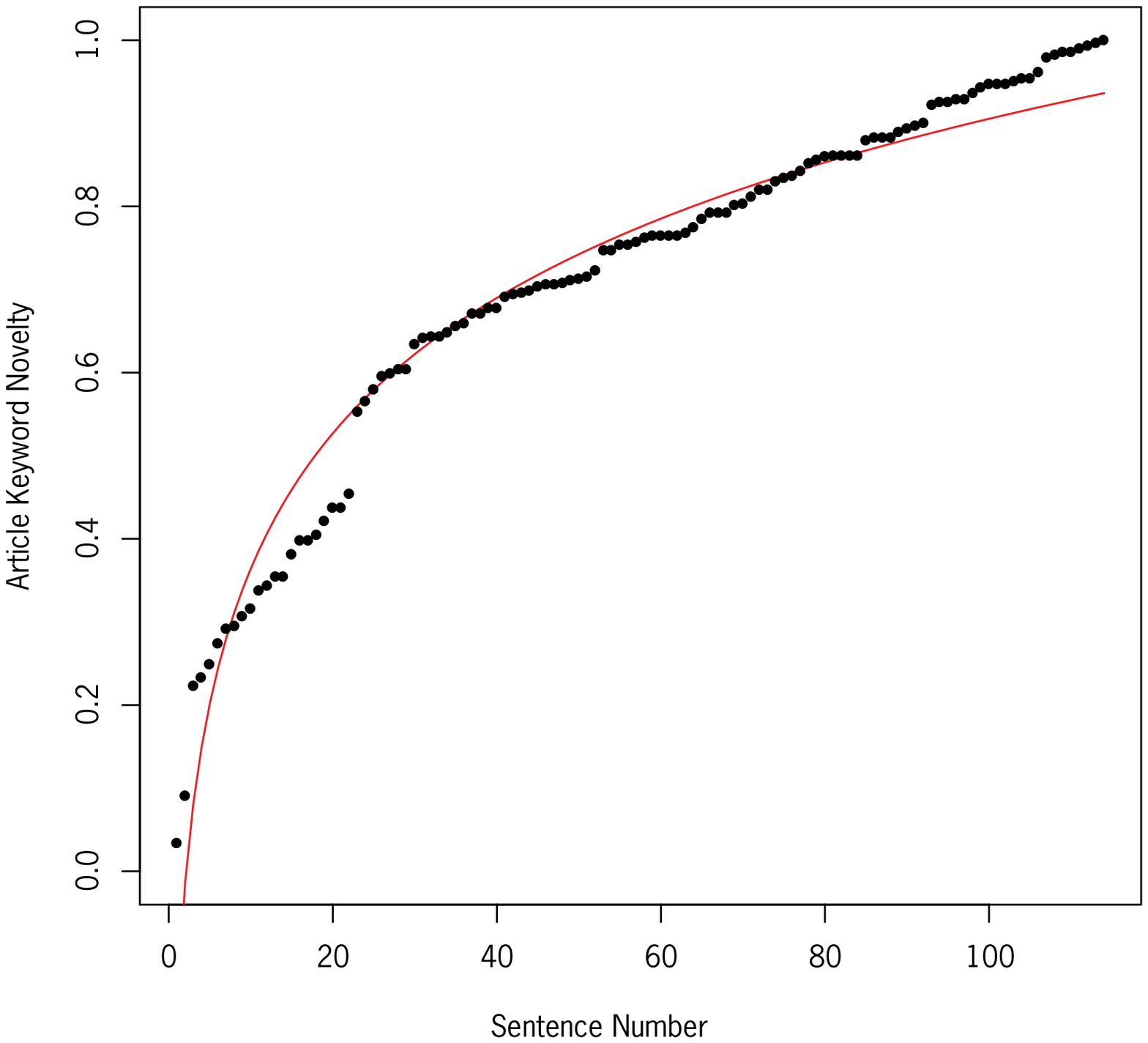}
 \caption{Technology}
 \label{fig:divMost}
 \end{subfigure}
 ~ 
 \begin{subfigure}[b]{\noveltySize\textwidth}
 \centering
 \includegraphics[width=\textwidth]{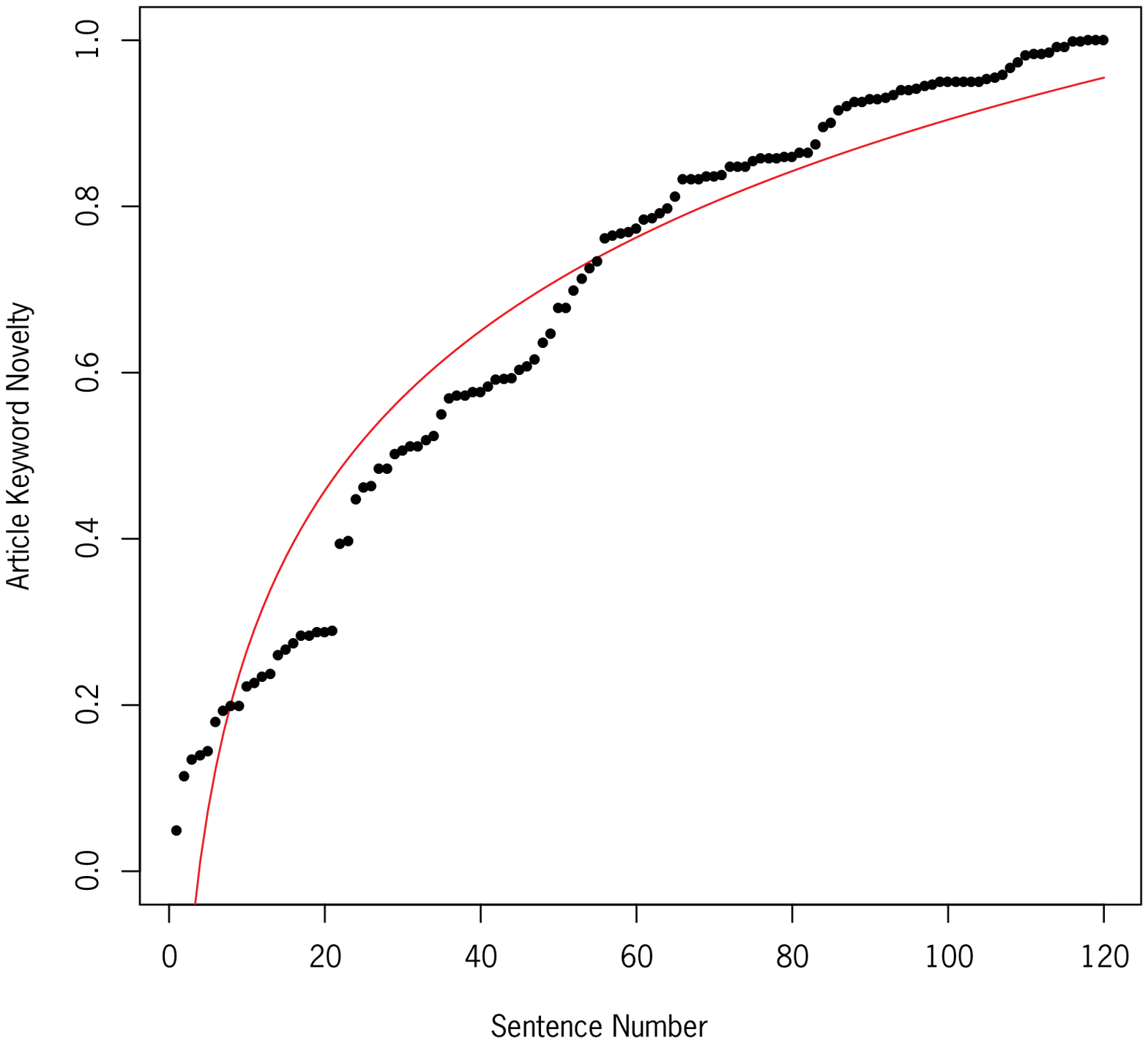}
 \caption{US News}
 \label{fig:divMost}
 \end{subfigure}
 ~ 
 \begin{subfigure}[b]{\noveltySize\textwidth}
 \centering
 \includegraphics[width=\textwidth]{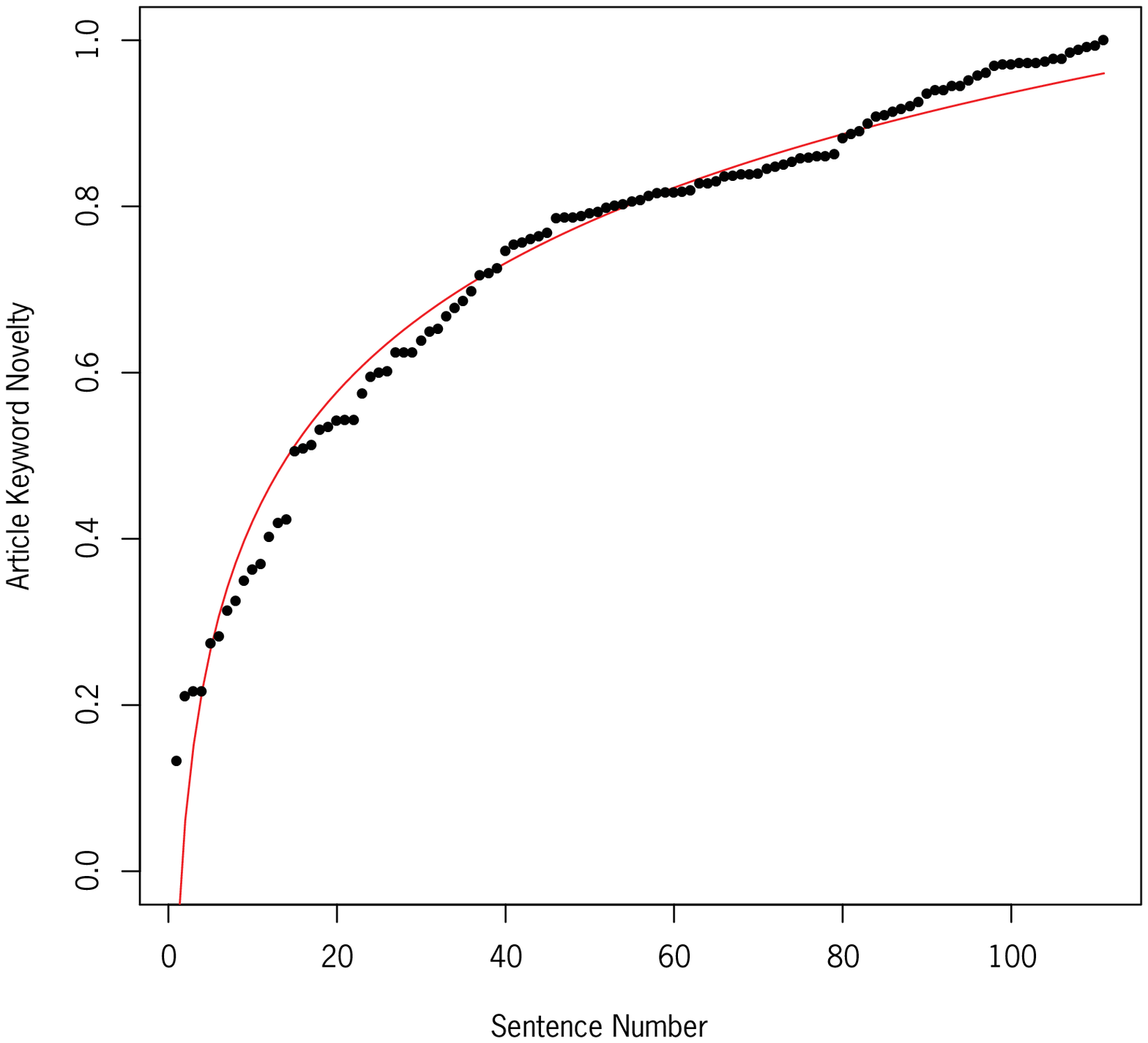}
 \caption{World News}
 \label{fig:divMost}
 \end{subfigure}
 \caption{Examples of Article Keyword Novelty Curves}{{\em {\scriptsize Points are presented with the resulting regression line in red | Graphs are high resolution best viewed in PDF}}}\label{fig:noveltyCurves}
\end{figure*}

%% file: slmFigure.tex
\begin{figure*}[!htb]
 \centering
  \begin{subfigure}[b]{\slmSize\textwidth}
 \centering
 \pgfplotsset{width=\textwidth,compat=1.3}
\begin{tikzpicture}
	\begin{axis}[
			xlabel={\slmXAxis}, 	xtick pos=left, 
			ytick pos=left,		ylabel={\slmYAxis},	ymax = .4,
			y tick label style={/pgf/number format/.cd, fixed, fixed zerofill, precision=2, /tikz/.cd},
			xticklabels/.append style={font=\smlFontSize},
			every axis/.append style={font=\smlFontSize},
		]
		\addplot [only marks, \slmPointsColor,mark size=\markSize] table [col sep=comma,x=SentenceIndex, y=kldArticle] {slm_graphs/buisness_78905.txt};
	\end{axis}
	\pgfplotsset{every axis y label/.append style={rotate=180}}
	\begin{axis}[
			ylabel={\slmYYAxis},		label style={\slmLineColor},	axis y line*=right,	ymax = \slmMaxY,
			hide x axis,
			every axis/.append style={font=\smlFontSize,\slmLineColor},
			extra y ticks = 2, 	extra y tick labels = ,		extra y tick style  = { grid = major,dashed },
		]
		\addplot [no markers, \slmLineColor] table [col sep=comma,x=SentenceIndex, y=sdAwayKldArticle] {slm_graphs/buisness_78905.txt};
	\end{axis}
\end{tikzpicture}
 \caption{Buisness}
 \label{fig:splash}
 \end{subfigure}%
 ~ 
   \begin{subfigure}[b]{\slmSize\textwidth}
 \centering
 \pgfplotsset{width=\textwidth,compat=1.3}
\begin{tikzpicture}
	\begin{axis}[
			xlabel={\slmXAxis}, 	xtick pos=left, 
			ytick pos=left,		ylabel={\slmYAxis},	ymax = .2,
			y tick label style={/pgf/number format/.cd, fixed, fixed zerofill, precision=2, /tikz/.cd},
			xticklabels/.append style={font=\smlFontSize},
			every axis/.append style={font=\smlFontSize},
		]
		\addplot [only marks, \slmPointsColor,mark size=\markSize] table [col sep=comma,x=SentenceIndex, y=kldArticle] {slm_graphs/ent_173355.txt};
	\end{axis}
	\pgfplotsset{every axis y label/.append style={rotate=180}}
	\begin{axis}[
			ylabel={\slmYYAxis},		label style={\slmLineColor},	axis y line*=right,	ymax = \slmMaxY,
			hide x axis,
			every axis/.append style={font=\smlFontSize,\slmLineColor},
			extra y ticks = 2, 	extra y tick labels = ,		extra y tick style  = { grid = major,dashed },
		]
		\addplot [no markers, \slmLineColor] table [col sep=comma,x=SentenceIndex, y=sdAwayKldArticle] {slm_graphs/ent_173355.txt};
	\end{axis}
\end{tikzpicture}
 \caption{Entertainment}
 \label{fig:slm_entertainment}
 \end{subfigure}%
 ~ 
   \begin{subfigure}[b]{\slmSize\textwidth}
 \centering
 \pgfplotsset{width=\textwidth,compat=1.3}
\begin{tikzpicture}
	\begin{axis}[
			xlabel={\slmXAxis}, 	xtick pos=left, 
			ytick pos=left,		ylabel={\slmYAxis},	ymax = 0.6,
			y tick label style={/pgf/number format/.cd, fixed, fixed zerofill, precision=2, /tikz/.cd},
			xticklabels/.append style={font=\smlFontSize},
			every axis/.append style={font=\smlFontSize},
		]
		\addplot [only marks, \slmPointsColor,mark size=\markSize] table [col sep=comma,x=SentenceIndex, y=kldArticle] {slm_graphs/pol_174073.txt};
	\end{axis}
	\pgfplotsset{every axis y label/.append style={rotate=180}}
	\begin{axis}[
			ylabel={\slmYYAxis},		label style={\slmLineColor},	axis y line*=right,	ymax = \slmMaxY,
			hide x axis,
			every axis/.append style={font=\smlFontSize,\slmLineColor},
			extra y ticks = 2, 	extra y tick labels = ,		extra y tick style  = { grid = major,dashed },
		]
		\addplot [no markers, \slmLineColor] table [col sep=comma,x=SentenceIndex, y=sdAwayKldArticle] {slm_graphs/pol_174073.txt};
	\end{axis}
\end{tikzpicture} \caption{Politics}
 \label{fig:slm_politics}
 \end{subfigure}%
 ~ 
   \begin{subfigure}[b]{\slmSize\textwidth}
 \centering
 \pgfplotsset{width=\textwidth,compat=1.3}
\begin{tikzpicture}
	\begin{axis}[
			xlabel={\slmXAxis}, 	xtick pos=left, 
			ytick pos=left,		ylabel={\slmYAxis},	ymax = .8,
			y tick label style={/pgf/number format/.cd, fixed, fixed zerofill, precision=2, /tikz/.cd},
			xticklabels/.append style={font=\smlFontSize},
			every axis/.append style={font=\smlFontSize},
		]
		\addplot [only marks, \slmPointsColor,mark size=\markSize] table [col sep=comma,x=SentenceIndex, y=kldArticle] {slm_graphs/sports_164428.txt};
	\end{axis}
	\pgfplotsset{every axis y label/.append style={rotate=180}}
	\begin{axis}[
			ylabel={\slmYYAxis},		label style={\slmLineColor},	axis y line*=right,	ymax = \slmMaxY,
			hide x axis,
			every axis/.append style={font=\smlFontSize,\slmLineColor},
			extra y ticks = 2, 	extra y tick labels = ,		extra y tick style  = { grid = major,dashed },
		]
		\addplot [no markers, \slmLineColor] table [col sep=comma,x=SentenceIndex, y=sdAwayKldArticle] {slm_graphs/sports_164428.txt};
	\end{axis}
\end{tikzpicture} \caption{Sports}
 \label{fig:slm_sports}
 \end{subfigure}%

   \begin{subfigure}[b]{\slmSize\textwidth}
 \centering
 \pgfplotsset{width=\textwidth,compat=1.3}
\begin{tikzpicture}
	\begin{axis}[
			xlabel={\slmXAxis}, 	xtick pos=left, 
			ytick pos=left,		ylabel={\slmYAxis},	ymax = .5,
			y tick label style={/pgf/number format/.cd, fixed, fixed zerofill, precision=2, /tikz/.cd},
			xticklabels/.append style={font=\smlFontSize},
			every axis/.append style={font=\smlFontSize},
			extra y ticks = 2, 	extra y tick labels = ,		extra y tick style  = { grid = major,dashed },
		]
		\addplot [only marks, \slmPointsColor,mark size=\markSize] table [col sep=comma,x=SentenceIndex, y=kldArticle] {slm_graphs/tech_150913.txt};
	\end{axis}
	\pgfplotsset{every axis y label/.append style={rotate=180}}
	\begin{axis}[
			ylabel={\slmYYAxis},		label style={\slmLineColor},	axis y line*=right,	ymax = \slmMaxY,
			hide x axis,
			every axis/.append style={font=\smlFontSize,\slmLineColor},
			extra y ticks = 2, 	extra y tick labels = ,		extra y tick style  = { grid = major,dashed },
		]
		\addplot [no markers, \slmLineColor] table [col sep=comma,x=SentenceIndex, y=sdAwayKldArticle] {slm_graphs/tech_150913.txt};
	\end{axis}
\end{tikzpicture} \caption{Technology}
 \label{fig:slm_tech}
 \end{subfigure}%
 ~ 
   \begin{subfigure}[b]{\slmSize\textwidth}
 \centering
 \pgfplotsset{width=\textwidth,compat=1.3}
\begin{tikzpicture}
	\begin{axis}[
			xlabel={\slmXAxis}, 	xtick pos=left, 
			ytick pos=left,		ylabel={\slmYAxis},	ymax = .5,
			y tick label style={/pgf/number format/.cd, fixed, fixed zerofill, precision=2, /tikz/.cd},
			xticklabels/.append style={font=\smlFontSize},
			every axis/.append style={font=\smlFontSize},
		]
		\addplot [only marks, \slmPointsColor,mark size=\markSize] table [col sep=comma,x=SentenceIndex, y=kldArticle] {slm_graphs/us_152248.txt};
	\end{axis}
	\pgfplotsset{every axis y label/.append style={rotate=180}}
	\begin{axis}[
			ylabel={\slmYYAxis},		label style={\slmLineColor},	axis y line*=right,	ymax = \slmMaxY,
			hide x axis,
			every axis/.append style={font=\smlFontSize,\slmLineColor},
			extra y ticks = 2, 	extra y tick labels = ,		extra y tick style  = { grid = major,dashed },
		]
		\addplot [no markers, \slmLineColor] table [col sep=comma,x=SentenceIndex, y=sdAwayKldArticle] {slm_graphs/us_152248.txt};
	\end{axis}
\end{tikzpicture} \caption{US News}
 \label{fig:slm_usNews}
 \end{subfigure}%
 ~ 
   \begin{subfigure}[b]{\slmSize\textwidth}
 \centering
 \pgfplotsset{width=\textwidth,compat=1.3}
\begin{tikzpicture}
	\begin{axis}[
			xlabel={\slmXAxis}, 	xtick pos=left, 
			ytick pos=left,		ylabel={\slmYAxis},	ymax = .4,
			y tick label style={/pgf/number format/.cd, fixed, fixed zerofill, precision=2, /tikz/.cd},
			xticklabels/.append style={font=\smlFontSize},
			every axis/.append style={font=\smlFontSize},
		]
		\addplot [only marks, \slmPointsColor,mark size=\markSize] table [col sep=comma,x=SentenceIndex, y=kldArticle] {slm_graphs/world_151023.txt};
	\end{axis}
	\pgfplotsset{every axis y label/.append style={rotate=180}}
	\begin{axis}[
			ylabel={\slmYYAxis},		label style={\slmLineColor},	axis y line*=right,	ymax = \slmMaxY,
			hide x axis,
			every axis/.append style={font=\smlFontSize,\slmLineColor},
			extra y ticks = 2, 	extra y tick labels = ,		extra y tick style  = { grid = major,dashed },
		]
		\addplot [no markers, \slmLineColor] table [col sep=comma,x=SentenceIndex, y=sdAwayKldArticle] {slm_graphs/world_151023.txt};
	\end{axis}
\end{tikzpicture} \caption{World News}
 \label{fig:slm_worldNews}
 \end{subfigure}
  \caption{Examples of SLM Article Plots + Std Deviation in Slope Plot}{{\em {\scriptsize KL Divergence are Black Points (left axis), Standard Deviations Away in Gray (right axis) | Graphs are high resolution best viewed in PDF}}}\label{fig:slmCurves}
\end{figure*}
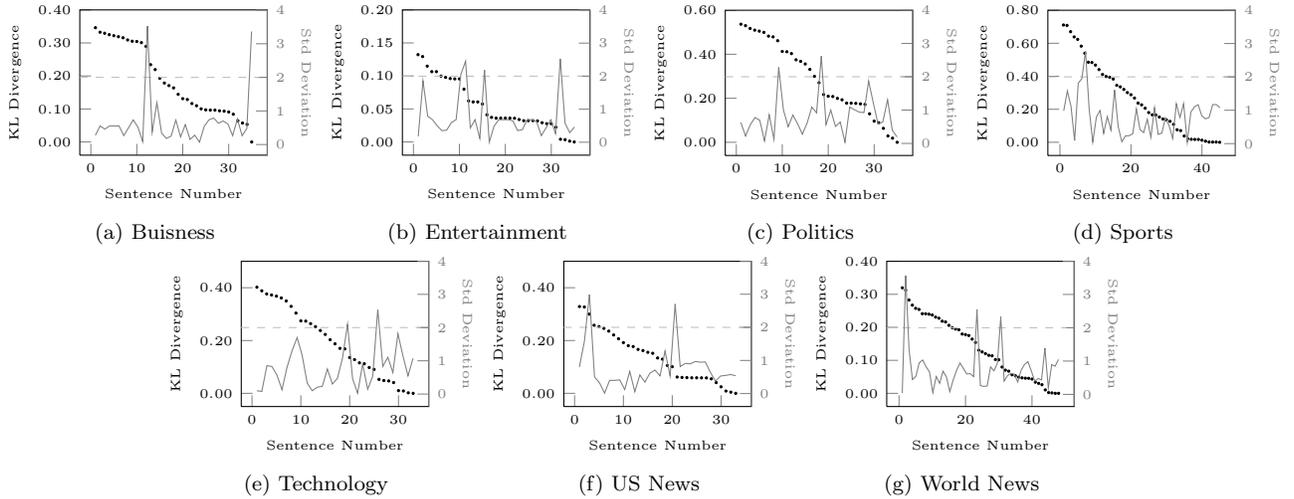

%% file: section_dataset.tex
\begin{table}
 \centering 
{\small
	\rowcolors{2}{tableShade}{white}
	\begin{tabular}{l  *{3}{c} }
	& {\bf Total} & {\bf Total} & {\bf $\mu$ Articles }\\
	{\bf Subject} & {\bf \# Corpuses} & {\bf \# Articles} & {\bf per Corpus (sd) }\\\toprule
	{\bf US}	&	183	&	19429	&	106.17 (52.81)\\
	{\bf Business}	&	166	&	19209	&	115.72 (63.45)\\
	{\bf Politics}	&	129	&	25094	&	194.53 (122.53)\\
	{\bf Entertainment}	&	218	&	30534	&	140.06 (78.70)\\
	{\bf World}	&	158	&	20440	&	129.37 (75.59)\\
	{\bf Sports	}&	203	&	27463	&	135.29 (65.08)\\
	{\bf Technology}	&	232	&	42104	&	181.48 (119.72)\\\hline
	{\bf All}	&	1289	&	184273	&	142.96 (90.73)

	\end{tabular}
}
\caption{Data Collected from Google News from 10/13 to 11/13}{{\scriptsize{\em  }}} 
\label{tab:fullDataSet}
\end{table}

\section{Master Dataset: Article Collection}

We collected corpuses of news articles by scraping Google News every 8 hours, beginning on October 21 to November 27 2013\footnote{Collection stopped when at least 100 corpuses per subject area were collected.}. Specifically, we collected corpuses from 7 subjects (World News, US News, Entertainment, Business, Technology, Sports and Politics).  Corpuses were `accepted' if they contained at least 50 unique articles. Google News therefore acted as a news aggregator, clustering the articles by topic. From the articles (grouped as a corpus by Google News), we retrieved the source HTML of the original article from the originally hosted website. The body copy of said article was then extracted using CETR \cite{Weninger:2010wn}. The description of the resulting articles and corpuses can be found in Table \ref{tab:fullDataSet}. To conduct each experiment, we randomly select articles and corpuses from this master dataset.

%% file: section_performance.tex
\section{Experiment: Model Performance} \label{sec:performance}

The primary experiment in this paper is to the evaluate ( in terms of break point semantic content) the performance of the 4 baseline lower bound metrics (One Sentence, Two Sentence, One Paragraph, and Twenty Percent) against the 4 lower bound candidate models (Article Keyword Novelty, Corpus Keyword Novelty, Article SLM, Corpus SLM). In this section we introduce the data set, then describe our methods used for comparing the above 8 algorithms for lower bound break point prediction.

\begin{table*}
 \centering 
{\small
	\rowcolors{2}{tableShade}{white}
	\begin{tabular}{l  *{6}{c} }
	{\bf }			&	{\bf Grade}	&	{\bf Reading}	&	{\bf Fog}		&	{\bf Sentence}	&	{\bf Word}			&	{\bf Corpus}\\
	{\bf }			&	{\bf Level}		&	{\bf Level}		&	{\bf Index}		&	{\bf Count}	&	{\bf Count}		&	{\bf Size}\\\toprule
	{\bf Business}	&	11.00 {\tiny(2.68)}	&	48.54 {\tiny(8.26)}	&	7.33 {\tiny(2.83)}	&	54.52 {\tiny(23.96)}	&	854.57 {\tiny(151.78)}	&	112.04 {\tiny(61.47)}	\\
	{\bf Ent.}		&	12.70 {\tiny(4.18)}	&	49.60 {\tiny(13.66)}	&	10.30 {\tiny(4.04)}	&	40.02 {\tiny(16.27)}	&	904 {\tiny(202.95)}		&	148.71 {\tiny(82.72)}	\\
	{\bf Politics}	&	13.16 {\tiny(3.03)}	&	43.64 {\tiny(10.77)}	&	9.72 {\tiny(3.02)}	&	40.15 {\tiny(13.61)}	&	887.93{\tiny(148.24)}		&	187.36 {\tiny(117.04)}	\\
	{\bf Sports}	&	11.00 {\tiny(7.15)}	&	58.54 {\tiny(19.28)}	&	9.55 {\tiny(7.39)}	&	42.11 {\tiny(13.51)}	&	875.98 {\tiny(176.31)}	&	127.89 {\tiny(60.35)}	\\
	{\bf Tech.}		&	12.23 {\tiny(4.37)}	&	49.55 {\tiny(14.91)}	&	9.53 {\tiny(4.19)}	&	42.31 {\tiny(15.20)}	&	903.67 {\tiny(197.98)}	&	170.9 {\tiny(109.85)}	\\
	{\bf US}		&	13.14 {\tiny(3.15)}	&	45.17 {\tiny(11.00)}	&	10.01 {\tiny(3.12)}	&	38.83 {\tiny(12.92)}	&	893.36 {\tiny(190.52)}	&	109.71 {\tiny(48.98)}	\\
	{\bf World}		&	13.97 {\tiny(3.13)}	&	40.32 {\tiny(11.83)}	&	10.27 {\tiny(2.95)}	&	38.93 {\tiny(12.57)}	&	925.18 {\tiny(190.72)}	&	130.38 {\tiny(72.28)}	\\\hline
	{\bf All}		&	12.46 {\tiny(4.32)}	&	47.91 {\tiny(14.23)}	&	9.53 {\tiny(4.30)}	&	42.41 {\tiny(16.61)}	&	892.10 {\tiny(184.32)}	&	141.00 {\tiny(86.56)}	
	\end{tabular}
}
 	 \caption{Documents Source Statistics}{{\scriptsize{\em Mean and Std Values Reported}}} 
 	 \label{tab:turkDocuments} 
 
 \end{table*}

\subsection{Dataset}

100 corpuses from each subject were randomly selected from the master dataset, with one article randomly selected from each corpus. This results in 100 articles from each subject, and 700 articles overall. This allows performance to be compared overall, and take into account the varying writing styles and language within each subject area. Summary statistics including Grade Level\footnote{Flesch-Kincaid grade level indicates that a student at that current U.S. school grade should be able to understand said document (e.g. 8.0 is eighth grade). 7.0 to 8.0 is `optimal.'}, Reading Level\footnote{The Flesch reading ease rates text on a 100 point scale, with higher scores being easier to understand. 60-70 is `optimal.'}, and Fog Index\footnote{Years of education to understand a document in a single reading (e.g. 12.0 is a high school senior). 8.0 is considered `optimal.'} are presented in Table  \ref{tab:turkDocuments}.

\subsection{Methods}

Amazon Mechanical Turk (MT) HITs were constructed from the 700 articles. A HIT is an individual task given to a person on MT.  Each HIT consisted of a brief definition of a break point and minimal break point, the original source text of one article\footnote{HITs were not grouped together (containing more than one article) and presentation order was random (reducing order effects).} and the 8 candidate break points presented in random order (reducing order effects). Before each sentence in the source text, we inserted the characters `(S{\em X})' where {\em X} was the sentence number. We used these indexes to refer to a given break point. For each break point, the participants were asked to respond to the statement {\em Is (SX) a good Minimal Break Point?} with a 7-point scale ({\em Too Short - Missing text }(1) to {\em Balanced }(4) to {\em Too Long - Extra Text }(7)). Each HIT was completed by 2 Master level Turkers\footnote{HIT Approval Rate above 95\% and at least 1000 approved HITS}, yielding 1400 measures of quality per model (100 articles across 7 areas).

To ensure `legitimate' HIT completion, one `sanity check question' was included asking Turkers to find the Nth word in the Mth sentence.  In addition, a HIT was rejected if the Turkers' response failed basic logic checks on their responses. First, all responses could not be the same. Second, if a single HIT asked about the same cut point more than once, that specific Turker's response must be the same to both questions (e.g. one Turker rating sentence \#4 as a 6 and then rating sentence \#4 as a 2 for the same article). Third, ratings must be in chronological order. For example, a sentence early in the document was listed as TOO LONG, and then a sentence later was listed as TOO SHORT, that doesn't make sense. However succeeding break points (in article order) could have the same ranking. If any of the three logic tests or the sanity check are failed, the HIT is rejected and re-posted.  Participants were remunerated 30 cents per accepted HIT.

An ANOVA and Student's T-test were used to compare the algorithms' performance. While performing multiple comparisons may suggest statistical adjustment to a more conservative value (i.e., Bonferroni correction), we chose to highlight multiple thresholds of significance following \cite{Savitz:1995un}. For transparency, we report results and summary statistics broken down by subject area. However, it is outside the scope of this paper to optimize for an individual subject area.

\subsubsection{Method Limitations}

Any evaluation with multiple comparisons has an interaction effect, in that the rating or quality of one break point can be impacted by the other break points offered for comparison. Thus, a break point that might have been rated a 5 (slightly too long) when viewed in isolation, may be pushed `higher' if there are other, less optimal break points located earlier in the article. Thus, these results must necessarily be viewed in the broader context of the other break points presented.

\input{./compareMethods}

\subsection{Results} 

Cross subject results are presented in Table \ref{tab:compareOverall} and results by subject area are in Table \ref{tab:compareBySubject}. ANOVA comparing the 8 algorithms resulted in a highly significant difference p$<$0.001 (F=1977). Pairwise comparative t-tests between algorithms were likewise all highly significant with with p$<$0.001, the one exception being the comparison between Two Sentences and One Paragraph, which had p=0.264. This similarity may be due to many articles breaking the first paragraph after two sentences. 

Overall, the most balanced approach (minimal extra text while retaining enough key content) was SLM Corpus followed closely by SLM Article. The three static baseline approaches (First Sentence, Second Sentence, First Paragraph) were all overly aggressive, cutting too much text and losing key information that was central to the article themselves. The remaining baseline approach (Twenty Percent) and the two Keyword Novelty methods were generally too relaxed with their break point prediction, choosing to break after superfluous text.

\subsubsection{Results: Correlations}

Given the varying readability and length of the articles (Table \ref{tab:turkDocuments}), we wanted to determine if any of those descriptive features influenced the performance of our algorithms. To this end, we tested the performances of each of the 8 algorithms against all readability statistics found in Table \ref{tab:turkDocuments} using a Spearman's rank test. No correlations were detected for any pairing of algorithm and readability statistic (with Spearman's rho never reaching above 0.2). This suggests that the performances reported in Table \ref{tab:compareOverall} \& \ref{tab:compareBySubject} were due to the algorithms themselves, and not the length or readability of the articles.

\input{./compareMethodsAll}

\subsection{Discussion: Algorithm Comparison}

Overall, the two SLM approaches performed quite well with a mean score of 4.18 and 4.66 for SLM-Corpus and SLM-Article respectively. The improvement in performance between the corpus and article versions is to be expected, since striving to reach the ideal language model that is the composite of an entire corpus of articles on a topic will smooth out author-specific phrasing, and focus more on the most central key words. However, it is not always realistic to assume that a layout system will have a broader corpus on which to build a more robust language model. Yet SLM Article, which built a language model only from the article, still outperforms the 3 other algorithms with scores above 5 (the two Keyword Novelty approaches, and Twenty Percent).

Given these results, we can also consider First Sentence, Second Sentence, and First Paragraph approaches to be extremely unreasonable for real-world use. When optimizing a document layout, missing key content (score less than 4) is worse than than including extra text (score at or above 4), since by leaving out critical information, the reader can be misinformed. 

Thus, using SLM objectively produces minimal break points that are more respectful of the semantic content of the article, preventing over-cutting and misleading the readership. In some situations, these break points are more aggressive than other approaches, potentially saving space while maintaing content readability. Yet overall, whether more or less content is needed for a minimal break point, SLM ensures content readability rather than form (layout) over function (readability). 

\input{./data/SLM_corpus}

\subsubsection{A Closer Look at SLM}

Based on the high performance of SLM Corpus, we wanted to determine if there was a group of breakpoints that were indicative of failure, or success. First, we classified the break point ratings as too short (1-3), balanced (4), and too long (5-7). So as to compare across articles (which have varying length), we divided break points into percentage of document intervals (0-5\%, 5-10\%, etc), and tabulated the number of ratings in each interval for a given class. To compare across the three classes, we must consider the tabulation as a percentage of the total ratings in said class. The results are plotted in Figure \ref{SLM_corpus}.

Upon viewing the curves for the three classes, it is worth noting that the `Too Short' and `Balanced' curves are almost identical. More specifically, 82\% of the failed, 69\% of the balanced, and 34\% of the too long occur at less than 15\% of the document. These results suggest why absolute approaches (e.g. Twenty percent) may not be aggressive enough.  Further, these results also indicate that there does not appear to be a universal lower threshold below which we are assured to be `too short.'  Thus, having adaptive algorithms that are dependent upon the semantic content of a document is critical to achieving quality pagination.

%% file: compareMethods.tex
\begin{table}[t]
 \centering 
	\rowcolors{2}{white}{tableShade}
	\begin{tabular}{c l c c}
	& {\bf Algorithm} & {\bf Mean (sd)} & {\bf Median} \\\toprule

\cellcolor{white}	&	\sOne	&	1.78 (0.97)	&	1.00	 \\
\cellcolor{white}	&	\sTwo	&	2.47 (1.18)	&	2.00	 \\
\cellcolor{white}	&	\pOne	&	2.44 (1.29)	&	2.00	 \\
\cellcolor{white}	&	\tw	&	5.19 (1.20)	&	5.00	 \\
\cellcolor{white}	&	\na	&	5.83 (1.14)	&	6.00	 \\
\cellcolor{white}	&	\nc	&	5.52 (1.19)	&	6.00	 \\
\cellcolor{white}	&	\ka	&	4.66 (1.76)	&	5.00	 \\
\multirow{-8}{*}[-0.1em]{\begin{sideways} \cellcolor{white} {\bf All Subject Areas}\end{sideways}}	&	\kc	&	4.18 (1.65)	&	4.00	 \\
	
	\end{tabular}

\caption{Overall Performance}{{\scriptsize{\em \sparklineFootnote  }}} 
\label{tab:compareOverall}
\end{table}

%% file: compareMethodsAll.tex
\begin{table}
 \centering
	\rowcolors{2}{white}{tableShade}
	\begin{tabular}{c l c c}
	& {\bf Algorithm} & {\bf Mean (sd)} & {\bf Median}  \\\midrule

\cellcolor{white}	&	\sOne	&	2.12 (1.08)	&	2.00	 \\
\cellcolor{white}	&	\sTwo	&	2.83 (1.32)	&	3.00	 \\
\cellcolor{white}	&	\pOne	&	2.78 (1.41)	&	3.00	 \\
\cellcolor{white}	&	\tw	&	5.33 (1.23)	&	5.00	 \\
\cellcolor{white}	&	\na	&	5.70 (1.11)	&	6.00	 \\
\cellcolor{white}	&	\nc	&	5.61 (1.17)	&	6.00	 \\
\cellcolor{white}	&	\ka	&	4.99 (1.81)	&	5.00	 \\
\multirow{-8}{*}[-0.1em]{\begin{sideways} \cellcolor{white} {\bf Buisness}\end{sideways}}	&	\kc	&	4.47 (1.64)	&	5.00	 \\
\hline\hline
\cellcolor{white}	&	\sOne	&	1.43 (0.72)	&	1.00	 \\
\cellcolor{white}	&	\sTwo	&	2.08 (0.92)	&	2.00	 \\
\cellcolor{white}	&	\pOne	&	2.24 (1.05)	&	2.00	 \\
\cellcolor{white}	&	\tw	&	5.23 (1.10)	&	5.00	 \\
\cellcolor{white}	&	\na	&	5.88 (1.03)	&	6.00	 \\
\cellcolor{white}	&	\nc	&	5.42 (1.05)	&	6.00	 \\
\cellcolor{white}	&	\ka	&	4.70 (1.78)	&	5.00	 \\
\multirow{-8}{*}[-0.1em]{\begin{sideways} \cellcolor{white} {\bf Entertainment}\end{sideways}}	&	\kc	&	3.82 (1.62)	&	3.00	 \\
\hline\hline
\cellcolor{white}	&	\sOne	&	1.72 (0.93)	&	1.00	 \\
\cellcolor{white}	&	\sTwo	&	2.43 (1.17)	&	2.00	 \\
\cellcolor{white}	&	\pOne	&	2.40 (1.31)	&	2.00	 \\
\cellcolor{white}	&	\tw	&	5.30 (0.91)	&	5.00	 \\
\cellcolor{white}	&	\na	&	5.89 (0.81)	&	6.00	 \\
\cellcolor{white}	&	\nc	&	5.65 (0.94)	&	6.00	 \\
\cellcolor{white}	&	\ka	&	4.79 (1.62)	&	5.00	 \\
\multirow{-8}{*}[-0.1em]{\begin{sideways} \cellcolor{white} {\bf Politics}\end{sideways}}	&	\kc	&	4.37 (1.48)	&	5.00	 \\
\hline\hline
\cellcolor{white}	&	\sOne	&	1.62 (0.86)	&	1.00	 \\
\cellcolor{white}	&	\sTwo	&	2.42 (1.06)	&	2.00	 \\
\cellcolor{white}	&	\pOne	&	2.44 (1.40)	&	2.00	 \\
\cellcolor{white}	&	\tw	&	5.40 (1.17)	&	6.00	 \\
\cellcolor{white}	&	\na	&	6.12 (1.02)	&	6.00	 \\
\cellcolor{white}	&	\nc	&	5.81 (1.10)	&	6.00	 \\
\cellcolor{white}	&	\ka	&	4.52 (1.65)	&	4.00	 \\
\multirow{-8}{*}[-0.1em]{\begin{sideways} \cellcolor{white} {\bf Sports}\end{sideways}}	&	\kc	&	4.40 (1.59)	&	4.00	 \\
\hline\hline
\cellcolor{white}	&	\sOne	&	1.82 (0.99)	&	1.00	 \\
\cellcolor{white}	&	\sTwo	&	2.50 (1.31)	&	2.00	 \\
\cellcolor{white}	&	\pOne	&	2.47 (1.27)	&	2.00	 \\
\cellcolor{white}	&	\tw	&	5.13 (1.33)	&	5.00	 \\
\cellcolor{white}	&	\na	&	5.78 (1.32)	&	6.00	 \\
\cellcolor{white}	&	\nc	&	5.27 (1.42)	&	6.00	 \\
\cellcolor{white}	&	\ka	&	4.58 (1.89)	&	5.00	 \\
\multirow{-8}{*}[-0.1em]{\begin{sideways} \cellcolor{white} {\bf Technology}\end{sideways}}	&	\kc	&	4.10 (1.83)	&	4.00	 \\
\hline\hline
\cellcolor{white}	&	\sOne	&	1.74 (0.94)	&	1.00	 \\
\cellcolor{white}	&	\sTwo	&	2.39 (1.13)	&	2.00	 \\
\cellcolor{white}	&	\pOne	&	2.27 (1.27)	&	2.00	 \\
\cellcolor{white}	&	\tw	&	5.05 (1.17)	&	5.00	 \\
\cellcolor{white}	&	\na	&	5.79 (1.15)	&	6.00	 \\
\cellcolor{white}	&	\nc	&	5.50 (1.19)	&	6.00	 \\
\cellcolor{white}	&	\ka	&	4.46 (1.81)	&	4.00	 \\
\multirow{-8}{*}[-0.1em]{\begin{sideways} \cellcolor{white} {\bf US News}\end{sideways}}	&	\kc	&	3.90 (1.61)	&	4.00	 \\
\hline\hline
\cellcolor{white}	&	\sOne	&	2.02 (1.07)	&	2.00	 \\
\cellcolor{white}	&	\sTwo	&	2.64 (1.17)	&	2.00	 \\
\cellcolor{white}	&	\pOne	&	2.49 (1.24)	&	2.00	 \\
\cellcolor{white}	&	\tw	&	4.88 (1.33)	&	5.00	 \\
\cellcolor{white}	&	\na	&	5.66 (1.39)	&	6.00	 \\
\cellcolor{white}	&	\nc	&	5.37 (1.32)	&	6.00	 \\
\cellcolor{white}	&	\ka	&	4.55 (1.71)	&	5.00	 \\
\multirow{-8}{*}[-0.1em]{\begin{sideways} \cellcolor{white} {\bf World News}\end{sideways}}	&	\kc	&	4.20 (1.67)	&	4.00	 \\
	
	\end{tabular}

\caption{Performance Per Subject}{{\scriptsize{\em  \sparklineFootnote}}} 
\label{tab:compareBySubject}
\end{table}

%% file: data/SLM_corpus.tex
\begin{figure}
 \centering
 \pgfplotsset{width=2.7in}
\begin{tikzpicture}
	\begin{axis}[
			xlabel={Everything Up To X\% of Document}, 	xtick pos=left, 
			ytick pos=left,		ylabel={Cumulative \% of Ratings},	
			y tick label style={/pgf/number format/.cd, fixed, fixed zerofill, precision=0, /tikz/.cd},
			xticklabels/.append style={font=\smlFontSize},
			every axis/.append style={font=\smlFontSize},
			legend pos=outer north east,
			axis lines=left,
			extra x ticks = {15,20}, 	extra x tick labels = ,		extra x tick style  = { grid = major, \slmLineColor },
		]
		\addplot [no markers, black ,mark size=\markSize] table [col sep=comma,x=lookup, y=TooShort] {data/SLM_corpus.csv};
		\addplot [no markers, black ,dashed] table [col sep=comma,x=lookup, y=Ideal] {data/SLM_corpus.csv};
		\addplot [no markers, black , thick] table [col sep=comma,x=lookup, y=TooLong] {data/SLM_corpus.csv};
		\legend{Too Short,Balanced, Too Long}
	\end{axis}
\end{tikzpicture} 
  \caption{When Failure Occurs}{{\em {\scriptsize Gray vertical lines at 15\% and 20\% for reference}}}\label{SLM_corpus}
\end{figure}

%% file: section_gold.tex
\section{Gold Standard Generation}

While the above work strove to identify an improved algorithm to better predict a lower bound break point based on semantics, we postulated that there may be a truly `ideal' break point in each article.  This `ideal' would be the natural place to break the article, and unlike the minimal break point, this ideal could be longer and contain more than just the bare minimum of information. To this end, we attempted to create a `gold standard' dataset. 

\subsection{Study Design}

The goal of this study was to find consensus on an ideal break point for a given article for the construction of a gold standard.  We followed the same random selection technique as described in Section \ref{sec:performance}, pulling 700 unique articles (100 from each subject area). We used CrowdFlower ({\bf CF}) as our crowd-sourced platform. Unlike MT, CF has a premium set of crowd-sourced individuals called the Editorial Crowd. From the CF website:  
{\quote
\vspace{-0.8em}
{\em This group of contributors have been tested for a deep understanding of the English language. These contributors have shown that they understand syntax, grammar, punctuation, and other elements of the English language.}
\vspace{-0.6em}

}
We realize that this task, attempting to find consensus, is a challenge and we therefore opted to use the more expensive CF verified workers with an expertise in English editing to ensure a careful reading and quality consideration of the ideal break point. We will refer to workers on CF as {\bf CFWs}.

CF HITs were constructed from the 700 randomly selected articles. A HIT consisted of a brief definition of a break point and the original source text (using the same sentence numbering as in the above experiment). The text was accompanied by the question, `What is the best break point in this document?'  CFWs were asked to choose one of the sentence numbers as the best break point (indicating everything before that number would be kept, and everything after would be cut). A full definition of an ``ideal'' break point was included to clear up any ambiguity on the labeling task. To ensure legitimate HIT completion, one `sanity check question' was included asking CFWs to find the Nth word in the Mth sentence. Failed sanity check question HITs were reposted. 

The CF infrastructure groups HITs together. Thus, each CFW was required to complete 4 HITs at a time, and was remunerated a total of 40 cents for those 4 HITs. Every HIT was evaluated by 5 unique CFWs, totaling 3500 observations. 

Given the subjective nature of these observations, we wished to only accept a break point (and its corresponding article) as a gold standard if a majority (at least 3 of the 5 CFWs) could agree upon an ideal break point, within a tolerance.

\begin{table}
 \centering 

	\rowcolors{3}{white}{tableShade}
	\begin{tabular}{l c c c c c}
	{\bf } & {\bf No } & {\bf 2 } & {\bf 3 } & {\bf 4 } & {\bf  5}\\
	{\bf Sentence} & {\bf CFW} & {\bf  CFW} & {\bf CFW } & {\bf  CFW} & {\bf CFW}\\
	{\bf Tolerance} & {\bf Agree} & {\bf Agree} & {\bf Agree} & {\bf Agree} & {\bf Agree}\\\hline
	 {\bf 0 Away}  & 361 & 292 & 46 & 1 & 0\\
	 {\bf 1 Away}  & 168 & 345 & 139 & 34 & 15\\
	 {\bf 2 Away}  & 92 & 304 & 201 & 68 & 35\\
	  {\bf 3 Away}  & 42 & 263 & 240 & 95 & 60\\
	\end{tabular}

\caption{Gold Standard Consensus for 700 Random Articles}{{\scriptsize{\em  Values represent the number of articles where consensus was at the column level given the row's tolerance. For example, there were 139 articles where 3 CFWs agreed on a break point within a 1 sentence tolerance (those three breakpoints were separated by no more than one sentence.)}}} 
\label{tab:gold}
\end{table}

\subsection{Results \& Discussion}

In a surprising turn of events, we saw an extremely low level of CFW agreement, as illustrated in Table \ref{tab:gold}. As part of our analysis, we include a range of agreement tolerance values (how many sentences are allowed to separate the break points chosen by k CFWs such that they could still be said to agree?)  With the strictest tolerance (break points must exactly match,), only 46 of the 700 articles had 3 out of 5 CFWs agreeing, 1 had 4 CFWs agreeing, and no articles had all 5 CFWS agreeing. This results in only 6.7\% of the articles reaching any semblance of consensus.  Even at a very generous 3 sentence tolerance (there may be up to 3 sentences between break points chosen by k CFWs, and agreement could still be claimed), only 395 articles (slightly above 50\%) reached a consensus at or above 3 CFWs. 

Given these low consensus results, the ability to meaningfully construct a gold standard is extremely limited. This suggests that the use of machine learning techniques, which rely upon a training set, would be prohibitive in this context.

Furthermore, this lack of consensus may suggest that there is no single `ideal' break point in a given article. Beyond finding a minimal `do not cut before this place' break point, the remaining text of the article (and its importance to the reader) may be too subjective to readily predict, and subsequently, this task may be ``too hard for a human.'' This adds further support to the importance and tractable outcome observed in our minimal break point detection.

%% file: section_conclusion.tex
\section{Future Work}

The first area of future work would be to develop a variation of PDM \cite{DameraVenkata:2011hv,Ahmadullin:2013dj} that integrates pagination quality as part of the layout quality metric. Currently these approaches are based on fit of content alone. Subsequently, integrating pagination quality would be a substantial challenge, but a critical next step to testing the impact of pagination break point quality in a existing print publications.

Second, we are examine the applications of pagination break points' application to e-documents. In this context, there is an explicit metric of quality to test layout impact: number of clicks of a ``read more'' link after the initial document content. This is another exciting application that would require a large amount of structural work to set up an experiment.

One last area of future work is to explore alternative statistical language models, including those that do not accord equal weight to each occurrence of a word. This could further improve SLM performance, especially for the SLM Article context.  We still believe that the Keyword Novelty approach has merit; however a more `aggressive' keyword weighting algorithm may be needed to cause the curves' inflection point to occur earlier in the article. 

\section{Conclusion}

Automatic document layout is rapidly becoming central to the production of most commercially printed newspapers and magazines. Within this context, pagination is a common layout challenge. However, all existing approaches to automatically calculate article break points for pagination neglect to account for the semantic content of the presented article. Disregarding the semantic information within the document can lead to overly aggressive or overly relaxed cutting, thereby running the risk of either eliminating key content and confusing the reader, or leaving in superfluous content and boring the reader, as well as making automatic layout more difficult.

We seek to directly address this shortcoming in this work. We present the first semantic-based pagination algorithm, to the authors' knowledge, for news article layout. Our approach predicts minimal break points based on the semantic content of an article through the use of a statistical language model. This approach is tested via a multi-subject experiment on 700 documents, comparing our method to 4 currently employed baselines and 3 alternative semantic approaches (also created for this paper). Results from this experiment clearly show that our approach strongly outperforms the baselines and alternatives. To further explore break point detection for pagination, we investigate whether an `ideal' break point could be found for paginating news articles. Results from a second study suggest that humans are not able to agree on a single best break point, suggesting that it is more practical to define reliable minimal break points for pagination tasks.

This work presents a strong validation that break point detection for pagination tasks can benefit from an examination of the semantic content of the news articles themselves. Within a real-world context, a combination of semantic-based lower bound break point prediction and spacial/aesthetic optimization is ideal for automated document synthesis.